\pdfoutput=1

\documentclass[11pt]{article}

\usepackage{ACL2023}

\usepackage{times}
\usepackage{latexsym}

\usepackage[T1]{fontenc}

\usepackage[utf8]{inputenc}

\usepackage{microtype}

\usepackage{inconsolata}

\usepackage{graphicx}
\usepackage{float}
\usepackage{enumitem}
\usepackage{verbatim}
\usepackage{hyperref}
\usepackage{blindtext}
\usepackage{booktabs}

\title{Replicable Benchmarking of Neural Machine Translation (NMT) on Low-Resource Local Languages in Indonesia}

\author{
Lucky Susanto$^{\S,\spadesuit}$ \\ lucky.susanto@ui.ac.id \And
Ryandito Diandaru$^{\S,\diamondsuit}$ \\ 13519157@std.stei.itb.ac.id \AND
Adila Krisnadhi$^{\spadesuit}$ \\ adila@cs.ui.ac.id  \And
Ayu Purwarianti$^{\diamondsuit}$ \\ ayu@itb.ac.id \And
Derry Wijaya$^{\clubsuit,\heartsuit}$ \\ derry.wijaya@monash.edu \AND
\normalfont{$^\spadesuit$University of Indonesia} \quad \normalfont{$^\diamondsuit$Bandung Institute of Technology}
\quad \normalfont{$^{\clubsuit}$Boston University} \AND
\quad \normalfont{$^{\heartsuit}$Monash University Indonesia} 
}

\begin{document}
\maketitle
\def\thefootnote{\S}\footnotetext{Equal contributions.}\def\thefootnote{\arabic{footnote}}

\begin{abstract}
Neural machine translation (NMT) for low-resource local languages in Indonesia faces significant challenges, including the need for a representative benchmark and limited data availability. This work addresses these challenges by comprehensively analyzing training NMT systems for four low-resource local languages in Indonesia: Javanese, Sundanese, Minangkabau, and Balinese. Our study encompasses various training approaches, paradigms, data sizes, and a preliminary study into using large language models for synthetic low-resource languages parallel data generation. We reveal specific trends and insights into practical strategies for low-resource language translation. Our research demonstrates that despite limited computational resources and textual data, several of our NMT systems achieve competitive performances, rivaling the translation quality of zero-shot gpt-3.5-turbo. These findings significantly advance NMT for low-resource languages, offering valuable guidance for researchers in similar contexts.
\end{abstract}

\section{Introduction}
Neural Machine Translation (NMT) holds a crucial role for local languages in Indonesia, supporting language documentation \cite{abney2010human}, native language preservation \cite{bird2012machine, nllb}, and bridging socioeconomic gaps \cite{socioeconomic-gaps}. However, challenges unique to low-resource languages have hindered progress in this field \cite{aji-etal-2022-one}. Our work addresses these challenges for four prominent local languages in Indonesia: Javanese, Sundanese, Minangkabau, and Balinese.

Impressive NMT advancements often come from well-resourced entities (i.e., Google's PaLM2 \cite{palm2}, OpenAI's gpt-3.5-turbo \cite{brownfewshot}, Facebook's NLLB-200 \cite{nllb}), focusing primarily on high-resource languages like English. This phenomenon highlights a research gap for languages with limited resources in data availability and computing power. For instance, benchmark NMT systems like NLLB-200 \cite{nllb} rely on substantial computing power, a luxury many researchers lack, especially those working with local Indonesian languages \cite{cahyawijaya2022nusacrowd}. This hampers progress due to the difficulty of gauging whether a new approach, method, architecture, or data augmentation would help improve model performance.

In this work, our contribution is a replicable benchmark of NMT systems for these local Indonesian languages trained on publicly available data and tested on the publicly available FLORES-200 dataset. We prioritize accessible computing resources. Our base cross-lingual \cite{lample2019crosslingual} XLM model uses only a modest compute setup. It is trained with only two languages at a time and on a single GPU with at most 48GB of memory, which we believe is within the reach of most researchers in this domain. We also only use publicly available data sources to train, including the NusaCrowd repository of Indonesian languages \cite{cahyawijaya2022nusacrowd} and parsed wikidumps results \footnote{\href{https://dumps.wikimedia.org/}{Wikidumps Page}}. Extending prior work, such as \cite{winata2023nusax}, we benchmark NMT models on multiple domains. 

In addition, in a preliminary study, we explore the impact of using gpt-3.5 \cite{brownfewshot} for synthetic low-resource language data generation to augment training. We also investigate code-switching's potential \cite{kuwanto2021lowresource} for improving low-resource language NMT that was previously unexplored for Indonesian languages.

\section{Related Work}

\subsection{NMT Benchmarks for Low-resource Local Languages in Indonesia}
Neural Machine Translation (NMT) benchmarks are pivotal in documenting and preserving low-resource local languages like those in Indonesia. Prior works \cite{nllb, cahyawijaya2022nusacrowd, winata2023nusax} have contributed to the creation of these NMT Benchmarks in two significant ways: (1) the creation and compilation of datasets accompanied by (2)  exploration and evaluation of different methodologies.

\citet{nllb} focused on developing NMT Benchmarks for low-resource languages. Their NLLB-200 model supports 200+ languages with more than 40K translation directions. Among these 200+ languages, some are local languages in Indonesia. They obtain state-of-the-art results for many translation directions through massive data collection efforts and computing resources. 

Similarly, \citet{winata2023nusax} also collects a multilingual dataset for both machine translation and sentiment analysis for ten local languages in Indonesia. They use the collected dataset to create a benchmark for both tasks, obtaining impressive results in the machine translation tasks for the review domain by fine-tuning pre-trained models. \citet{winata2023nusax} also shows that fine-tuning a non-English-centric pre-trained model on local Indonesian languages outperforms its English-centric counterpart for the machine translation task. 


However, it is essential to recognize that the NMT benchmark created by \citet{nllb} is challenging to replicate. Many researchers and institutions, including top Indonesian universities \cite{cahyawijaya2022nusacrowd}, do not have access to massive compute resources or extensive and proprietary training data required to train the NLLB-200 models. Meanwhile, the NMT benchmark created by \citet{winata2023nusax} is limited only to the review domain. These leave a research gap that needs to be filled by benchmark NMT models that are replicable and cover more general domains. 

\subsection{High-resource vs Low-resource NMT}
Unlike low-resource NMT systems (where either the source or target language is a low-resource language), NMT systems for high-resource languages have achieved impressive results \cite{nllb}. Even with the progress achieved by the grassroots movement mentioned in the previous section, the performance gaps are wide. This phenomenon is due to research in the field of NMT and NLP being dominated by English and other major languages, which means that more efforts have significantly been put into developing language technologies for these significant languages, and more data have been collected and made available for these languages \cite{akhbardeh-etal-2021-findings-wmt21, kocmi-etal-2022-findings-wmt22}. This means that while low-resource NMTs face problems no longer found in high-resource NMTs, insufficient resources and attention are being allocated.

One significant issue NMT systems face is the pivotal role parallel data plays in model performance \cite{koehn-knowles-2017-sixchallenges}. By definition, low-resource languages have little to no parallel data. One problem that negatively impacts the model performance is out-of-vocabulary (OOV) occurrences \cite{aji-etal-2022-one,wibowo-etal-2021-indocollex}, where the model needs to see a token more to learn what it means. While this issue exists even in high-resource languages, the rate of occurrence for low-resource languages is substantially higher. However, usage of byte pair encoding (BPE) \cite{sennrich2016neural} is capable of alleviating this issue to some degree \cite{lample2018phrasebased, yang-etal-2020-csp}.

While previous research has made noteworthy strides in addressing out-of-vocabulary (OOV) occurrences, the most effective solution continues to be expanding available training data. However, building textual resources for translation tasks necessitates a significant investment of money, time, and expertise. Because of this, current research increasingly centers on finding innovative ways to augment the training of NMT models.

\subsection{Augmenting Training for NMT}
To combat the issue of data starvation, many researchers aim to utilize monolingual data to train NMT systems \cite{lample2018unsupmt, artetxe2018unsupervised, lample2019crosslingual} and find ways to generate more training data, either comparable or synthetic data. Comparable data are extracted using various bitext retrieval methods \cite{zhao2002adaptive,fan2021beyond,jones2021majority,kocyigit2022better}, multimodal signals \cite{hewitt2018learning,rasooli-etal-2021-wikily}, dictionary- or knowledge-based approaches \cite{wijaya2016mapping,wijaya2017learning,tang2022knowledge}; while synthetic data are created and utilized either through innovative training data augmentation \cite{kuwanto2021lowresource}, utilizing automatic back-translation \cite{sennrich2016improving, wang-etal-2019-english}, or even outright generating synthetic data using generative models \cite{lu-2023-syntheticdata}, which has gained increasing attention by the community lately due to the advancement of large language models (LLMs).

Both \citet{artetxe2018unsupervised} and \citet{lample2018unsupmt} show that NMT systems can be trained using only monolingual data while achieving impressive results. \citet{lample2019crosslingual} then create the XLM architecture, which allows NMT systems to be traine using monolingual and parallel data. Afterward, \cite{kuwanto2021lowresource} exploits the cross-lingual nature of the XLM models by corrupting the monolingual data using code-switching, which makes a single training instance contain multiple languages. The result is an improvement in the model's performance for low-resource translation.

In addition, prior works also focus on obtaining synthetic training data by turning monolingual data into parallel data through automatic back-translation \cite{sennrich2016improving} or by using LLMs such as the gpt-family models \cite{brownfewshot} that have been gaining popularity in recent years in many fields \cite{lu-2023-syntheticdata}. While back-translation has evolved, becoming a prominent method in the field of NMT \cite{artetxe2018unsupervised, lample2019crosslingual}, using LLMs to generate synthetic data has yet to be thoroughly explored. This trend of using generative AI to generate synthetic training data displays initial potential, considering their remarkable performances compared to the state-of-the-art in machine translation \cite{zhu2023-llmtranslator}. However, further research with ablation studies and the inclusion of more language coverage is still needed.

\section{Methodology}

In this section, we outline our methodology for creating a replicable NMT benchmark for four Indonesian languages: Javanese (jv), Sundanese (su), Minangkabau (min), and Balinese (ban). We aim to systematically explore different training approaches and paradigms for NMT while maintaining a consistent base architecture (XLM), fixed hyperparameters, and controlled computing environment. Our compute environment is given a strict upper bound, in which a total of 48 GPU Hours from a single GPU for each model training, totaling up to 96 GPU Hours for NMT systems utilizing pre-trained language models. We also limit the memory of the GPU used to a maximum of 48 GB. 

\subsection{Training Approaches}
We employ three primary training approaches to build our NMT models:

\textbf{From Scratch (Scratch)}: In this approach, models are trained from the ground up without any reliance on pre-existing pre-trained language models. This approach acts as a baseline and allows us to gauge the performance of the models when trained from scratch. 

\textbf{Pre-trained Cross-Lingual (PreXL)}: Here, an NMT model utilizes a pre-trained cross-lingual model (XLM) \cite{lample2019crosslingual} on two sets of monolingual data. One of the sets is the Indonesian monolingual data, and the other is the low-resource local language monolingual data. This provides a strong starting point for the NMT by initializing the model with knowledge from the target and source languages. Therefore, each language pair in this work is given its own respective model. The number of pre-trained models for \textbf{PreXL} equals the number of language pairs in our work, which is four.

\textbf{Code-switched Pre-trained Cross-Lingual (CodeXL)}: This approach involves pre-training the language model using additional augmented data from the two sets of monolingual data and a bilingual dictionary through code-switching, explained later in section \ref{code-switch-section}. Code-switching allows for a bilingual context within each training instance. The pre-trained model is then fine-tuned for translation. The tasks used to fine-tune \textbf{CodeXL} and \textbf{PreXL} depend on the training paradigm used (section \ref{paradigm-section}. The number of pre-trained models for \textbf{CodeXL} is the same as \textbf{PreXL}, which is four.

We chose the XLM architecture due to its modest compute resource requirements and its capability of cross-lingual language modeling. Moreover, the architecture is widely used for many low-resource language pairs and shows impressive results despite its modest size \cite{wang-etal-2019-english}. We use Masked Language Modeling (MLM) \cite{devlin2019bert} to pre-train all the XLM models.

\subsection{Training Paradigms}
\label{paradigm-section}

Additionally, we also explore two training paradigms, each influencing how the NMT models learn and what data are used for training:

\textbf{Unsupervised NMT (Unsup)}: This paradigm trains the NMT system using only monolingual data of the source and target language. Utilizing both denoising-autoencoding \cite{denoising-autoencoding} and automatic back-translation \cite{sennrich2016improving} to train the NMT system. Note that even though \textbf{CodeXL} utilizes a bilingual dictionary, it does not use any parallel data during pre-training. 

\textbf{Semi-supervised NMT (Semisup)}: This paradigm trains the NMT system using both monolingual data and parallel data of the source and target language. Monolingual data are utilized for training NMT by automatic back-translation.

We employ these shortened terms throughout our experiments to refer to the respective training approaches and paradigms. Our results indicate that the performance of each combination of approaches and paradigms on the evaluation dataset depends heavily on the amount of available data for the language: \textbf{Unsup} paradigm works better for very low-resource languages. In contrast, \textbf{Semisup} paradigm performs better when at least 10K parallel data is available \cite{artetxe2018unsupervised}. We do not conduct training using a strictly \textbf{Supervised NMT} paradigm because prior work has shown automatic back-translation's undeniable impact in improving low-resource NMT systems performance \cite{sennrich2016improving}.

\subsection{Training with Synthetic Data}

Following recent trends of using generative AI to generate synthetic training data \cite{lu-2023-syntheticdata, zhu2023-llmtranslator}, we explore the impact of synthetically generated data on low-resource language NMT systems. We define two main approaches to generating synthetic data: (1) generating parallel data using generative AI and (2) translating monolingual data using an existing model.

To gauge the impact of the synthetically generated training data, we train NMT systems with these additional data using the \textbf{Scratch} and \textbf{CodeXL} training approaches. \textbf{Scratch} is also used in our preliminary experiments to identify the synthetic data generation approach that would yield the best empirical results. Once we identify the best approach, we apply the same synthetic data generation approach to all our language pairs and use the generated data to augment the training of our NMT approach with the \textbf{Semisup} paradigm.

Through the preliminary experiments (reported in Appendix \ref{sec:appendix-C}), we find that synthetic data generated using generative AI (gpt-3.5-turbo) has the most positive impact on training NMT systems. We generate 5000 parallel sentences for each language pair via a zero-shot prompt\footnote{\href{https://github.com/Exqrch/IndonesianNMT/tree/master}{Repository} of the data we generate using zero-shot prompting}: \emph{\textbf{"Generate a long parallel sentence in SRC and TGT"}}, where \textbf{SRC} and \textbf{TGT} is the pair of language we want to generate the sentences in. Appendix \ref{sec:appendix-C} provides justifications for these choices. 


\subsection{Fine-tuning Objectives}
Denoising autoencoding \textbf{(DAE)} \cite{denoising-autoencoding} is a popular training objective for fine-tuning pre-trained LM for unsupervised MT tasks \cite{lample2018phrasebased, wang-etal-2019-english} for its ability to increase the robustness of NMT models.

By utilizing the XLM architecture, our NMT system can perform multi-way translation. Thus, we also utilize automatic back translation \textbf{(BT)} \cite{sennrich2016improving} during fine-tuning of our NMT models with \textbf{Unsup} and \textbf{Semisup} paradigms. By performing back translation using the same model that is being trained, synthetic parallel data is obtained and used automatically during training.

\subsection{Training Data}
\begin{table}[ht]
\small
    \centering
    \begin{tabular}{lrr}
    \toprule
        \textbf{Lang} & \textbf{Mono}  & \textbf{Para} \\
    \midrule
        \verb|jv| & \verb|1.6M| & \verb|14.3K|\\
        \verb|su| & \verb|550K| & \verb|13.2K|\\
        \verb|min| & \verb|282K| & \verb|17.2K|\\
        \verb|ban| & \verb|60K| & \verb|0.9K|\\
    \bottomrule
    \end{tabular}
    \caption{{Total number of monolingual sentences (\textbf{Mono}) and parallel sentences paired with Indonesian (\textbf{Para}) per language}}
    \label{tab:language-counts}
\end{table}

We obtain our monolingual data from multiple publicly available sources. For Indonesian (id), we use the 201M monolingual sentences available from the Indo4B curated dataset \cite{wilie2020indonlu}. We obtain monolingual data for the local languages through publicly available data such as Wikidumps\footnote{\href{https://dumps.wikimedia.org/}{Wikidumps Page}}, cc100 \cite{conneau-etal-2020-unsupervised}, imdb-jv \cite{wongso-2021}, jadi-ide \cite{jadi-ide}, and su-emot \cite{su-emot}. The amount of monolingual sentences used to train each language is available in Table \ref{tab:language-counts}, with further breakdown available in Appendix \ref{sec:appendix-A}.

All parallel data we use to train the model are also publicly available from the NusaCrowd repository \cite{cahyawijaya2022nusacrowd}. We scan the repository for datasets that contain parallel data of the local language paired with Indonesian. The amount of parallel sentences used to train each language pair is available in Table \ref{tab:language-counts}. Our largest language in terms of monolingual and parallel sentences, Javanese, is a tiny fraction (almost a 20th and a 500th, respectively) of NLLB-200 reported sentences for Javanese. From publicly available resources in Table \ref{tab:language-counts}, we can see that these four languages represent low-resource languages. A further breakdown is available in Appendix \ref{sec:appendix-B}.

The sentence counts in Table \ref{tab:language-counts} are after we perform filtering on both monolingual and parallel data. For monolingual data, we remove sentences that contain less than three words or more than 250 words. We also perform simple filtering for sentences obtained from Wikipedia, including deduplication, removing HTML tags, removing sentences with only numbers, removing sentences that do not start with an alphabet, and removing metadata, bulletin points, or number ordering from sentences. For parallel data, we remove sentences that contain less than three words or more than 250 words and remove sentence pairs whose source sentences have a word count ratio above 1.5 of their translations following the setup of \citet{ghazvininejad-2023}.

\subsection{Code Switch}
\label{code-switch-section}
\begin{figure}[ht]
    \centering
    \includegraphics[width=0.5\textwidth]{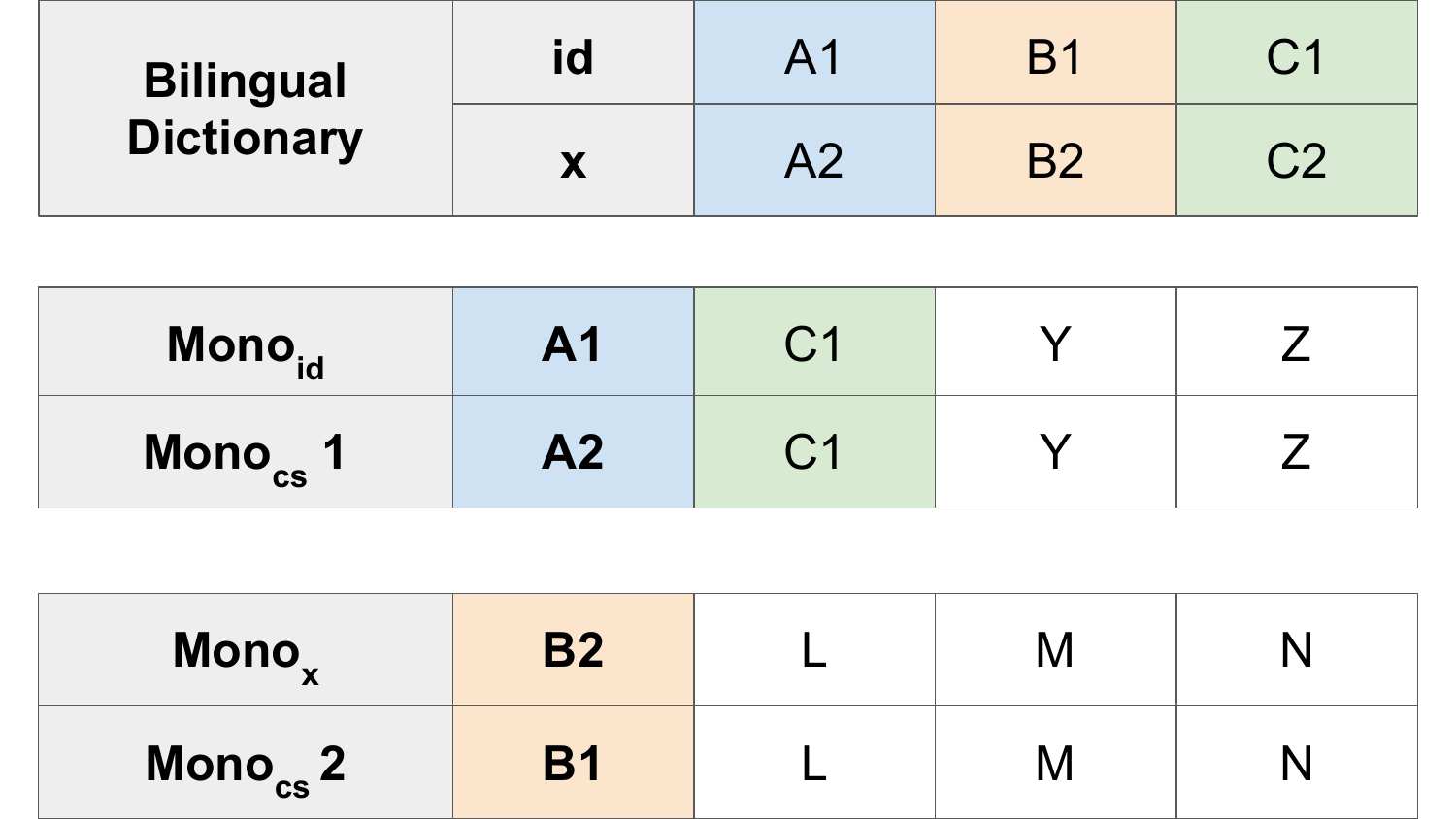}
    \caption{Illustration of generating synthetic data using code-switch. Using a bilingual dictionary, some words in each monolingual sentence are translated (i.e., blue and orange words). The candidate word for translation is chosen randomly, and not all words will be translated (green box).}  \label{fig:cs_illustration}
\end{figure}

In this paper, code-switching is done by utilizing the system made by \citet{kuwanto2021lowresource}\footnote{\href{https://gkuwanto.github.io/frontend/}{Code Switch-based Curriculum Training}}. Code-switching is used to create synthetic training data by utilizing a bilingual dictionary. The generated data is used only during pre-training and is treated as a third language (labeled \textbf{cs}), where each training instance contains tokens from the other two languages. Results obtained by \citet{kuwanto2021lowresource} imply that this method helps the model by giving stronger cross-lingual signals, which helps translation tasks during fine-tuning. 

\begin{table}[ht]
\small
    \centering
    \begin{tabular}{lrrr}
    \toprule
        \textbf{Lang} & \textbf{Mono$_{id}$} & \textbf{Mono$_{x}$} & \textbf{Mono$_{cs}$}\\
    \midrule
        \verb|{id, jv, cs}| & \verb|201M| & \verb|1.6M| & \verb|209M|\\
        \verb|{id, su, cs}| & \verb|201M| & \verb|550K| & \verb|208M|\\
        \verb|{id, min, cs}| & \verb|201M| & \verb|282K| & \verb|208M|\\
        \verb|{id, ban, cs}| & \verb|201M| & \verb|60K| & \verb|207M|\\
    \bottomrule
    \end{tabular}
    \caption{{Total number of monolingual training instances for each language in each NMT system. \textbf{Mono$_{id}$}, \textbf{Mono$_{x}$}, \textbf{Mono$_{cs}$} denotes the size of training instance for Indonesian, regional, and the third language, respectively.}}
    \label{tab:cs-data}
\end{table}
Creating synthetic training data through code-switching utilizes training data from monolingual datasets from both languages in the system. By utilizing a bilingual dictionary, obtained and parsed from \citet{winata2023nusax}\footnote{\href{https://github.com/IndoNLP/nusax/tree/main/datasets/lexicon}{NusaX's bilingual dictionary}}, each instance of training data from both monolingual datasets is augmented. Figure \ref{fig:cs_illustration} illustrates this process, while Table \ref{tab:cs-data} shows how much augmented training data is available for each NMT system.

\section{Experiment Results}
\begin{table*}[ht]
\small
\centering
    \begin{tabular}{llcccccc}
        \toprule
        \textbf{Translation} & \textbf{Paradigm} & \textbf{Turbo} & \textbf{Scratch} & \textbf{Scratch$_{AUG}$} & \textbf{PreXL} & \textbf{CodeXL} & \textbf{CodeXL$_{AUG}$}\\
        \midrule
        \verb|id-->jv| &Zero shot& 18.91 & --.-- & --.-- & --.-- & --.-- & --.--\\
        \verb|       | & Unsup     & --.-- & 12.06 & --.-- & 16.94 & \textit{17.90} & --.--\\  
        \verb|       | & Semisup   & --.-- & 13.50 & 18.29 & 19.31 & \textbf{\textit{21.32}} & 21.18\\  
        \midrule
        \verb|jv-->id| &Zero shot& \textbf{29.97} & --.-- & --.-- & --.-- & --.-- & --.--\\
        \verb|       | & Unsup   & --.-- & 08.11 & --.-- & 13.73 & \textit{20.93} & --.--\\   
        \verb|       | & Semisup & --.-- & 12.18 & 21.36 & 18.88 & 26.17 & \textit{26.23}\\ 
        \midrule
        \verb|id-->su| &Zero shot& 16.39 & --.-- & --.-- & --.-- & --.-- & --.--\\
        \verb|       | & Unsup   & --.-- & 10.42 & --.-- & \textit{14.69} & 10.68 & --.--\\  
        \verb|       | & Semisup & --.-- & 13.19 & 15.33 & 16.22 & \textbf{\textit{18.91}} & 18.90\\  
        \midrule
        \verb|su-->id| &Zero shot& \textbf{30.71} & --.-- & --.-- & --.-- & --.-- & --.--\\
        \verb|       | & Unsup   & --.-- & 08.07 & --.-- & \textit{13.62} & 10.68 & --.--\\   
        \verb|       | & Semisup & --.-- & 13.55 & 20.72 & 21.68 & 28.06 & \textit{28.40}\\ 
        \midrule
        \verb|id-->min| &Zero shot& 13.71 & --.-- & --.-- & --.-- & --.-- & --.--\\
        \verb|        | & Unsup   & --.-- & 10.18 & --.-- & 16.03 & \textit{18.37} & --.--\\  
        \verb|        | & Semisup & --.-- & 22.33 & 22.59 & 23.83 & \textbf{\textit{26.04}} & 25.18\\  
        \midrule
        \verb|min-->id| &Zero shot& 28.27 & --.-- & --.-- & --.-- & --.-- & --.--\\
        \verb|        | & Unsup   & --.-- & 07.88 & --.-- & 12.00 & \textit{19.43} & --.--\\   
        \verb|        | & Semisup & --.-- & 17.48 & 25.27 & 20.93 & 29.83 & \textbf{\textit{30.06}}\\ 
        \midrule
        \verb|id-->ban| &Zero shot& \textbf{14.94} & --.-- & --.-- & --.-- & --.-- & --.--\\
        \verb|       |  & Unsup   & --.-- & 08.28 & --.-- & 11.03 & \textit{12.70} & --.--\\  
        \verb|       |  & Semisup & --.-- & 00.22 & 00.30 & 02.63 & 06.36 & \textit{09.51}\\  
        \midrule
        \verb|ban-->id| &Zero shot& \textbf{26.93} & --.-- & --.-- & --.-- & --.-- & --.--\\
        \verb|        | & Unsup   & --.-- & 07.13 & --.-- & 10.34 & \textit{18.05} & --.--\\   
        \verb|        | & Semisup & --.-- & 00.30 & 00.36 & 05.35 & 11.16 & \textit{17.34}\\ 
        \bottomrule
    \end{tabular}
    \caption{
    \label{tab:all-v1}
        {Performance of NMT systems for Indonesian (\textbf{id}), Javanese (\textbf{jv}), Sundanese (\textbf{su}), Minangkabau (\textbf{min}), and Balinese (\textbf{ban}) translations. \textbf{Turbo} refers to gpt-3.5-turbo's zero-shot translation performance. \textbf{Zero shot} paradigm indicates translation without training. \textbf{AUG} denotes models trained with synthetic data generated by gpt-3.5-turbo. Bold values represent the best overall performance, while italicized values indicate the best performance within each paradigm.}
    }
\end{table*}
We concentrate our efforts on four language pairs: \textbf{id-jv}, \textbf{id-su}, \textbf{id-min}, and \textbf{id-ban}. Indonesia (\textbf{id}) is spoken by approximately 198 million people worldwide, whereas Javanese (\textbf{jv}), Sundanese (\textbf{su}), Minangkabau (\textbf{min}), and Balinese (\textbf{ban}) are spoken by roughly 68.2 million, 32.4 million, 4.8 million, and 3.3 million people, respectively, according to \cite{ethnologue2023}. Unsurprisingly, monolingual and parallel data availability for these four local languages in Indonesia generally follows a similar pattern. Javanese boasts the most extensive corpus of monolingual text data, while Balinese has the smallest. Regarding parallel data, Sundanese leads the way, closely followed by Javanese, while Balinese trails behind. Due to the substantial variation in training data availability, we present our findings for each local language separately. This approach allows us to assess the impact of different training methods and paradigms while assessing the influence of training data size.

We conduct experiments using three training approaches (\textbf{Scratch}, \textbf{PreXL}, \textbf{CodeXL}) and two training paradigms (\textbf{Unsup}, \textbf{Semisup}). For each combination of these approaches and paradigms, four NMT systems are trained (one of each language pair mentioned above). In total, there are 24 different NMT systems trained this way. 


In addition, we conduct experiments using synthetic parallel datasets. We only generate parallel training data, so these data do not affect the \textbf{Unsup} training paradigm. To evaluate the impact of these synthetic datasets, we employ two distinct training approaches: \textbf{Scratch} and \textbf{CodeXL}. We denote the process of training NMT systems with additional synthetic parallel training data as \textbf{Scratch$_{AUG}$} and \textbf{CodeXL$_{AUG}$}, respectively. These comprise the remaining 8 NMT systems created in this work, totaling 32. Since we use gpt-3.5-turbo to generate our synthetic parallel data in a zero-shot manner, we also benchmark the zero-shot translation performance of gpt-3.5-turbo (\textbf{Turbo}) on our evaluation dataset. 

The results of these experiments (all metrics are in spm200BLEU, shown in Table \ref{tab:all-v1}), reveal a consistent trend: \textbf{CodeXL} approach results in a significantly better performing NMT systems compared to \textbf{Scratch} and \textbf{PreXL}. An exception to this pattern is noted in the \textbf{id-su} language pair when employing the \textbf{Unsup} training paradigm.

\subsection{Javanese}
The \textbf{id-jv} language pair is particularly significant due to its relevance in Indonesia, where approximately 198 million people speak Indonesian (id), and Javanese (jv) is spoken by roughly 68.2 million \cite{ethnologue2023}. 

Looking at Table \ref{tab:all-v1}, we observe a substantial gap in translation performance between \textbf{id$\rightarrow$jv} and \textbf{jv$\rightarrow$id}, emphasizing the performance asymmetry. Notably, when training NMT systems using the \textbf{Unsup} paradigm, \textbf{CodeXL} consistently outperforms other approaches for both translation directions, reinforcing the findings of \citet{kuwanto2021lowresource}, which highlight the generalization capability of this approach.

Surprisingly, when we train NMT models using additional parallel data generated by gpt-3.5-turbo (\textbf{CodeXL$_{AUG}$}), we notice a slight decline in performance for \textbf{id$\rightarrow$jv} translation compared to our best-performing model (\textbf{CodeXL}). A more detailed comparison is discussed in the next section. 


\subsection{Sundanese}
Table \ref{tab:all-v1} shows where \textbf{id-su} differs from the \textbf{id-jv} language pair. For \textbf{id-su} language pair, the \textbf{Unsup} paradigm shows a different trend where \textbf{CodeXL} has a slightly worse performance compared to \textbf{PreXL} in the \textbf{id$\rightarrow$su} translation. However, this trend shifts when using the \textbf{Semisup} paradigm, with \textbf{CodeXL} regaining its superiority. 

Similar to \textbf{id-jv} language pair, an intriguing phenomenon arises when we train NMT models using additional parallel data generated by gpt-3.5-turbo (\textbf{CodeXL$_{AUG}$}) for the \textbf{id-su} language pair. While this approach does not create a better performing model in \textbf{id$\rightarrow$su} translation, it does result in a slightly better model for \textbf{su$\rightarrow$id}. This trend indicates that the generated synthetic parallel data's impact heavily depends on the generative AI's translation performance. For both \textbf{id-jv} and \textbf{id-su} language pairs, gpt-3.5-turbo's zero-shot translation performance on \textbf{id$\rightarrow$x} is worse than \textbf{CodeXL} for each respective language pair, therefore \textbf{CodeXL$_{AUG}$} does not result in improved performance. Meanwhile, the reverse is true, gpt-3.5-turbo's \textbf{x$\rightarrow$id} translation performance is better than \textbf{CodeXL} in \textbf{x$\rightarrow$id} direction, hence  \textbf{CodeXL$_{AUG}$} has a better performance in this direction. 

\subsection{Minangkabau}
The results we obtained for \textbf{id-min} follow a similar pattern as \textbf{id-jv}. The trend where \textbf{CodeXL} models performed better than \textbf{Scratch} and \textbf{PreXL} continues for \textbf{id-min} for both translation directions. 

However, unlike \textbf{id-jv} and \textbf{id-su}, using synthetically generated parallel data to train NMT systems for \textbf{id-min} (\textbf{CodeXL$_{AUG}$}) performed better than \textbf{CodeXL} on the \textbf{min$\rightarrow$id} translation. This is surprising because \textbf{CodeXL} performed better than \textbf{Turbo} on \textbf{min$\rightarrow$id}, yet the parallel data generated by \textbf{Turbo} was able to create \textbf{CodeXL$_{AUG}$}, which is a better performing NMT system. This breaks the previous trends set by \textbf{id-jv}, \textbf{id-su}, and even \textbf{id-ban} in the later section.


\subsection{Balinese}
\textbf{id-ban} continues the trend set by the majority of previous language pairs. Following \textbf{id-jv} and \textbf{id-min} language pairs, \textbf{CodeXL} consistently has superior performance compared to \textbf{Scratch} and \textbf{PreXL}.

Additionally, \textbf{id-ban} follows the trend set by \textbf{id-jv} and \textbf{id-su}, where the use of synthetically generated parallel data from \textbf{Turbo} creates a better NMT system compared to others that do not use them. For \textbf{id-ban} language pair specifically, \textbf{Turbo}'s translation performance is much higher than \textbf{CodeXL}, and the data \textbf{Turbo} generated has a significant impact during training, as seen in \textbf{CodeXL$_{AUG}$}. The difference in score for \textbf{CodeXL} and \textbf{CodeXL$_{AUG}$} differs by \textbf{3+} and \textbf{6+} spm200BLEU for \textbf{id$\rightarrow$ban} and \textbf{ban$\rightarrow$id} respectively. This performance difference is much more significant compared to \textbf{id-jv}, \textbf{id-su}, and \textbf{id-min}, where the performance difference is less than \textbf{1} spm200BLEU. This finding supports the idea that the generated synthetic parallel data's impact heavily depends on the generative AI's translation performance. Moreover, if the initial parallel data is limited, like in the case of \textbf{id-ban} (only 0.9K), the addition of synthetic data can greatly improve performance. 

However, Table \ref{tab:all-v1} shows that \textbf{Unsup} training paradigm created the best performing NMT system for \textbf{id-ban} language pair. While it would not be surprising for \textbf{Scratch} due to the limited amount of parallel training data, it is surprising that \textbf{Scratch$_{AUG}$} does not result in a considerably better NMT system, as the parallel training data size becomes 6x its original size (i.e., from 0.9K to 5.9K). This indicates that \textbf{denoising-autoencoding} plays a more significant role in model performance than parallel data when the training parallel data is limited. 



\section{Conclusion}
In this work, we create a replicable NMT benchmark under low-resource settings. We comprehensively train and analyze NMT systems for four low-resource Indonesian local languages: Javanese, Sundanese, Minangkabau, and Balinese. Our experiments shed light on the impact of different training approaches, paradigms, data sizes, and generated synthetic parallel data in low-resource local languages in Indonesia. In conclusion:

We observe that the \textbf{CodeXL} training approaches generally create NMT systems with better performances compared to \textbf{Scratch} and \textbf{PreXL} approaches. This further strengthens the robustness of the approach suggested by \citet{kuwanto2021lowresource}, where code-switching is used to give a stronger cross-lingual signal during model pre-training. Code-switching more positively impacts translation performance for \textbf{x$\rightarrow$id} more than \textbf{id$\rightarrow$x}. For reference, NMT systems created using the \textbf{CodeXL} training approach and \textbf{Semisup} paradigm have an average performance of \textbf{23.80} and \textbf{18.15} spm200BLEU for \textbf{x$\rightarrow$id} and \textbf{id$\rightarrow$x} respectively.

Furthermore, even after pre-training a language model using the MLM objective, fine-tuning the model using the denoising autoencoding objective might play a more prominent role in extremely low-resource NMT than just training the model to be more robust for the translation task. This is shown in the \textbf{id-ban} language pair NMT systems, where \textbf{Unsup} created better NMT systems than the \textbf{Semisup} training paradigm. It is noteworthy that the addition of 5000 synthetic parallel training data might not be enough to significantly improve NMT system performance, as visible in the \textbf{CodeXL}-\textbf{Unsup} entry compared to \textbf{CodeXL$_{AUG}$}-\textbf{Semisup} entry in Table \ref{tab:all-v1}, since the resulting parallel data is still very limited (i.e., less than 6K sentences). 

Lastly, we also observe a trend in which generative AIs can help augment the training process by generating synthetic parallel data. In most cases, excluding the \textbf{id-min} language pair, the parallel data generated by generative AI can impact the performance of NMT systems to approach or even outperform the performance of the generative AI's translation performance with much less compute and data resource. 

\section{Future Work}
Along with the above conclusions, our work also opens several venues for future research. Further ablation studies are needed to fully understand the impact of denoising-autoencoding on translation tasks. Our results indicate that the denoising-autoencoding objective not only increases model robustness but may also play a role in cross-lingual language understanding in extremely low-resource NMT.


In addition, further investigations into synthetically generated parallel data quality and diversity are crucial. We observe a trend where synthetically generated parallel data from gpt-3.5-turbo impact the training of NMT systems such that its performance approaches or even outperforms gpt-3.5-turbo's zero-shot translation performance. 

\section{Limitation}
While our work has given insight into NMT systems for low-resource local languages in Indonesia, it is essential to note that we have utilized different GPUs (TitanV, RTX8000, RTX6000, A6000, A40) with a maximum memory capacity of 48GB for different experiments. These GPU architectures and memory capacity variations may have influenced the observed performance. However, it is crucial to recognize that hardware differences alone cannot fully account for all the performance gaps observed. Future research should conduct experiments using a more standardized GPU setup to understand the impact of hardware variations better. 

Additionally, all of our experiments that include the utilization of gpt-3.5-turbo are problematic as it is a closed-sourced model. This causes problems such as transparency and reproducibility in the future. Future work should continue performing ablation studies on open-sourced LLMs. 

\section{Acknowledgement}
Authors from Indonesia are supported by the MoECRT ACE Open Research program. The authors also thank the Indonesian government for their funding and Boston University for providing essential computing resources. We also thanks Garry Kuwanto from Boston University for his help in utilizing his Code-switching system. 

\bibliography{anthology,custom}
\bibliographystyle{acl_natbib}

\newpage

\appendix

\onecolumn
\section{Monolingual Training Data Breakdown}
\label{sec:appendix-A}
\begin{table}[ht]
\small
    \centering
    \label{tab:daerah-source-mono}
    \begin{tabular}{lrrrrrrr}
    \hline
        \textbf{Lang} & \textbf{wikidumps}  & \textbf{cc100} & \textbf{imdb-jv} & \textbf{jadi-ide} & \textbf{su-emot} & \textbf{Total} & \textbf{Filter} \\
    \hline
        \verb|jv| & \verb|382K| & \verb|1.41M| & \verb|100K| & \verb|16K| & \verb|0| & \verb|1.91M| & \verb|1.6M|\\
        \verb|su| & \verb|220K| & \verb|387K|  & \verb|0| & \verb|0| & \verb|2K| & \verb|610K| & \verb|550K|\\
        \verb|min| & \verb|282K| & \verb|0|& \verb|0| & \verb|0| & \verb|0| & \verb|282K| & \verb|282K|\\
        \verb|ban| & \verb|60K| & \verb|0|  & \verb|0| & \verb|0| & \verb|0| & \verb|60K| & \verb|60K|\\
    \hline
    \end{tabular}
    \caption{Monolingual training data breakdown. \textbf{Lang} denotes the language identifier, \textbf{Total} denotes the total monolingual sentences for each language, and \textbf{Filter} denotes how many monolingual sentences remain after filtering, as listed in our Methodology section.}
    \label{tab:app-a}
\end{table}

The monolingual data of local languages in Indonesia are obtained from multiple sources, including parsed wikidumps, cc100 \cite{conneau-etal-2020-unsupervised}, imdb-jv \cite{wongso-2021}, jadi-ide \cite{jadi-ide}, and su-emot \cite{su-emot}. Excluding wikidumps monolingual data, which was taken in December of 2022, all of these sources are obtained from the compilation done by \citet{cahyawijaya2022nusacrowd} and was taken in January of 2023. The breakdown for these monolingual data of local languages in Indonesia is found in Table \ref{tab:app-a}

For the monolingual data of the Indonesian language, we use the Indo4B curated dataset \cite{wilie2020indonlu}. Excluding the data obtained from Twitter, the number of monolingual Indonesian sentences is 201 million. No sentences were filtered out due to the high quality of the dataset.

\section{Parallel Training Data Breakdown}
\label{sec:appendix-B}
\begin{table}[ht]
\small
    \centering
    \label{tab:daerah-source-para}
    \begin{tabular}{lrrrrrrrr}
    \hline
        \textbf{Lang} & \textbf{su-id}  & \textbf{min-nlp} & \textbf{code-mixed} & \textbf{bible} & \textbf{nusantara} & \textbf{nusax} & \textbf{Total} & \textbf{Filter}\\
    \hline
        \verb|jv| & \verb|0| & \verb|0| & \verb|977| & \verb|7958| & \verb|6000| & \verb|1000| & \verb|15935| & \verb|14395|\\
        \verb|su| & \verb|3616| & \verb|0|  & \verb|0| & \verb|7957| & \verb|1699| & \verb|1000| & \verb|14272| & \verb|13269|\\
        \verb|min| & \verb|0| & \verb|16371|& \verb|0| & \verb|0| & \verb|0| & \verb|1000| & \verb|17371| & \verb|17260|\\
        \verb|ban| & \verb|0| & \verb|0|  & \verb|0| & \verb|0| & \verb|0| & \verb|1000| & \verb|1000| & \verb|997|\\
    \hline
    \end{tabular}
    \caption{Parallel training data breakdown of the language pair Indonesia and the local language denoted by \textbf{Lang} (i.e., the entry \textbf{jv} list how much parallel data of the pair \textbf{ind, jv} are obtained from each source). \textbf{Total} denotes total parallel sentences for each pair of Indonesian and local languages, and \textbf{Filter} denotes how much remains after filtering, as listed in our Methodology section.}
    \label{tab:app-b}
\end{table}

As with our monolingual data breakdown, all of our parallel data were obtained from the NusaCrowd repository in January 2023. The datasets we use include su-id \cite{su-id}, min-nlp \cite{min-nlp}, code-mixed \cite{code-mixed}, bible \cite{cahyawijaya-etal-2021-indonlg}, nusantara \cite{korpus-nusantara}, and nusax \cite{winata2023nusax}.

\section{Ablation of Different Methods in Generating Parallel Data}
\label{sec:appendix-C}

\begin{table}[ht]
\small
    \centering
    \begin{tabular}{lrr}
    \hline
        \textbf{Model} & \textbf{id$\rightarrow$jv}  & \textbf{jv$\rightarrow$id} \\
    \hline
        \verb|text-davinci-003| & \verb|17.95| & \verb|26.26|\\
        \verb|gpt-3.5-turbo|    & \verb|19.10| & \verb|30.19|\\
    \hline
    \end{tabular}
    \caption{Zero-shot translation spm200BLEU score of generative AIs on the FLORES200 test set. Results indicate that gpt-3.5-turbo performs significantly more than text-davinci-003 on zero-shot translation.}
    \label{tab:llm-idjv}
\end{table}

We performed exploratory experiments regarding different methods of generating parallel data. As mentioned in our methodology, we define two main approaches to generating synthetic data: (1) Generating parallel data using generative AI and (2) Translating monolingual data using an already trained model.

\begin{table}[ht]
\small
    \centering
    \begin{tabular}{lccc}
    \hline
        \textbf{Model} & \textbf{Approach} & \textbf{id$\rightarrow$jv}  & \textbf{jv$\rightarrow$id} \\
    \hline
        \verb|Baseline|         & -         & \verb|12.99| & \verb|12.29|\\
        \hline
        \verb|text-davinci-003| & zero-shot & \verb|14.86| & \verb|16.88|\\
        \verb||                 & ten-shot  & \verb|14.40| & \verb|16.82|\\   
        \hline
        \verb|gpt-3.5-turbo|    & zero-shot & \verb|14.95| & \textbf{17.10}\\
        \verb||                 & ten-shot  & \textbf{15.04} & \verb|16.87|\\
    \hline
    \end{tabular}
    \caption{Performance of NMT systems trained from scratch using a \textbf{Semisup} paradigm when the original parallel data is mixed with synthetically generated data from gpt-3.5-turbo or text-davinci-003. The first row is the baseline model performance. All BLEU scores are from XLM's automated BLEU scoring. Bolded entries indicate the model with the best performance.}
    \label{tab:n4n6vx}
\end{table}

The models we use to generate the parallel data in approach (1) are gpt-3.5-turbo and davinci-text-003. We limit our exploratory experiment to the \textbf{id$\leftrightarrow$jv} translation direction. First, we compare the zero-shot translation performance of these models on the FLORES200 test set, where gpt-3.5-turbo achieved a considerably higher spm200BLEU score. The full breakdown is available in Table \ref{tab:llm-idjv}. We give each model the prompt to generate these parallel data: \textbf{"Generate a long parallel sentence in SRC and TGT"}. Our internal experiments show that without the keyword \textbf{"long"}, the model will generate short and simple parallel sentences consisting of regularly occurring words.

\begin{figure}[ht]
    \centering
    \includegraphics[width=1\textwidth]{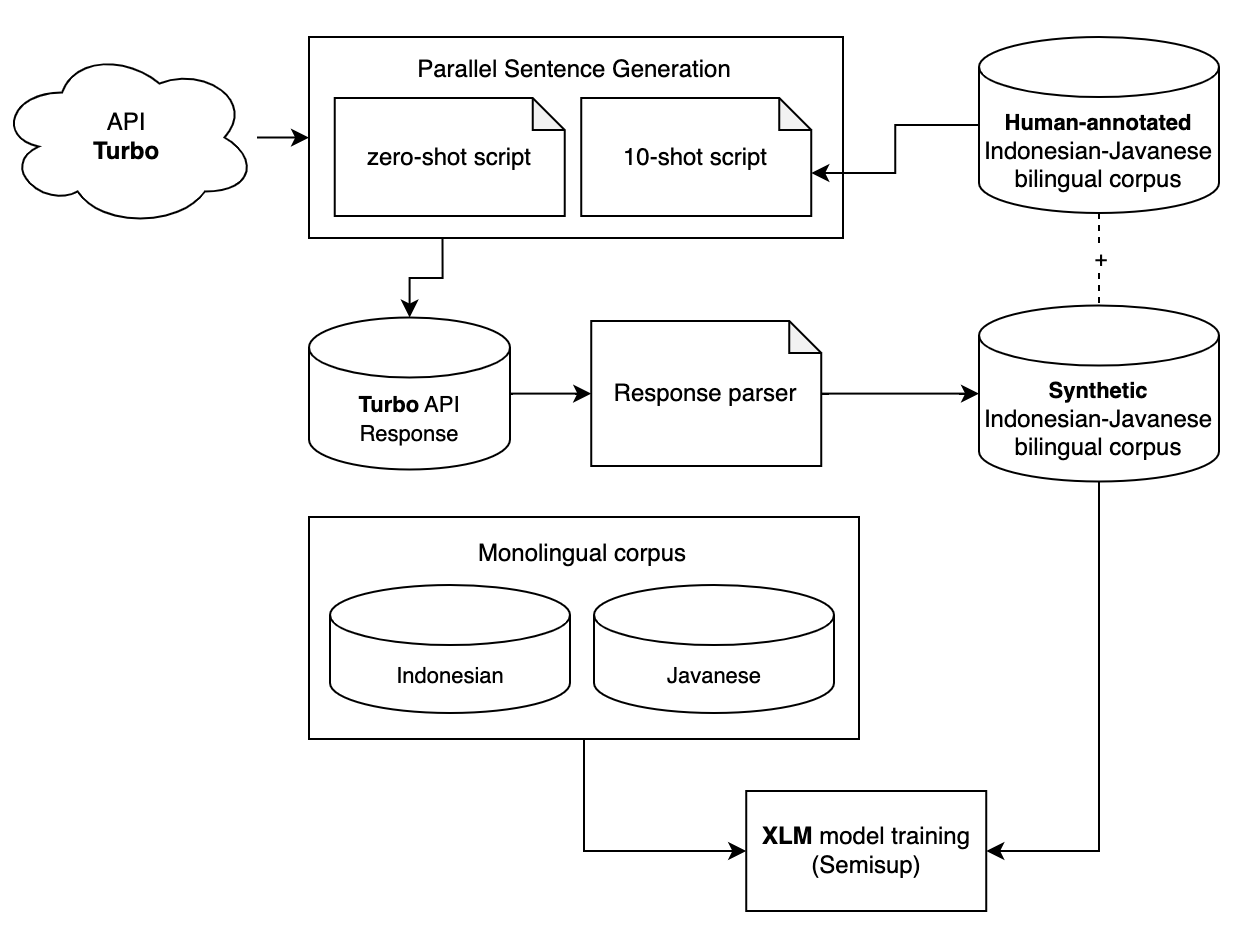}
    \caption{Illustration of using LLMs to generate synthetic parallel data. First tested on \textbf{id-jv} language pair, we use the same pipeline to generate synthetic parallel data for the other language pair.}
    \label{fig:generation_illustration}
\end{figure}

We conduct these experiments using two approaches: zero-shot generation and ten-shot generation. We give the model the prompt above without additional context in zero-shot generation. We then parse the text that has been generated and split it into Indonesian sentences and Javanese sentences. In the ten-shot generation, we sample 10 parallel sentences in our original training data to feed it as examples to the model. Figure \ref{fig:generation_illustration} illustrates this process. The impact these generated synthetic data have on training is found in Table \ref{tab:n4n6vx}. These performances align with the benchmark results in Table \ref{tab:llm-idjv}, where gpt-3.5-turbo is better at both translation directions than text-davinci-003. The results in Table \ref{tab:n4n6vx} show that zero-shot generation of gpt-3.5-turbo creates parallel data with the most positive impact on NMT system training. The results shown in Table \ref{tab:n4n6vx} indicate that few-shot may not have a considerable difference in performance compared to zero-shot for LLMs on translation tasks.

\begin{figure}[ht]
    \centering
    \includegraphics[width=1\textwidth]{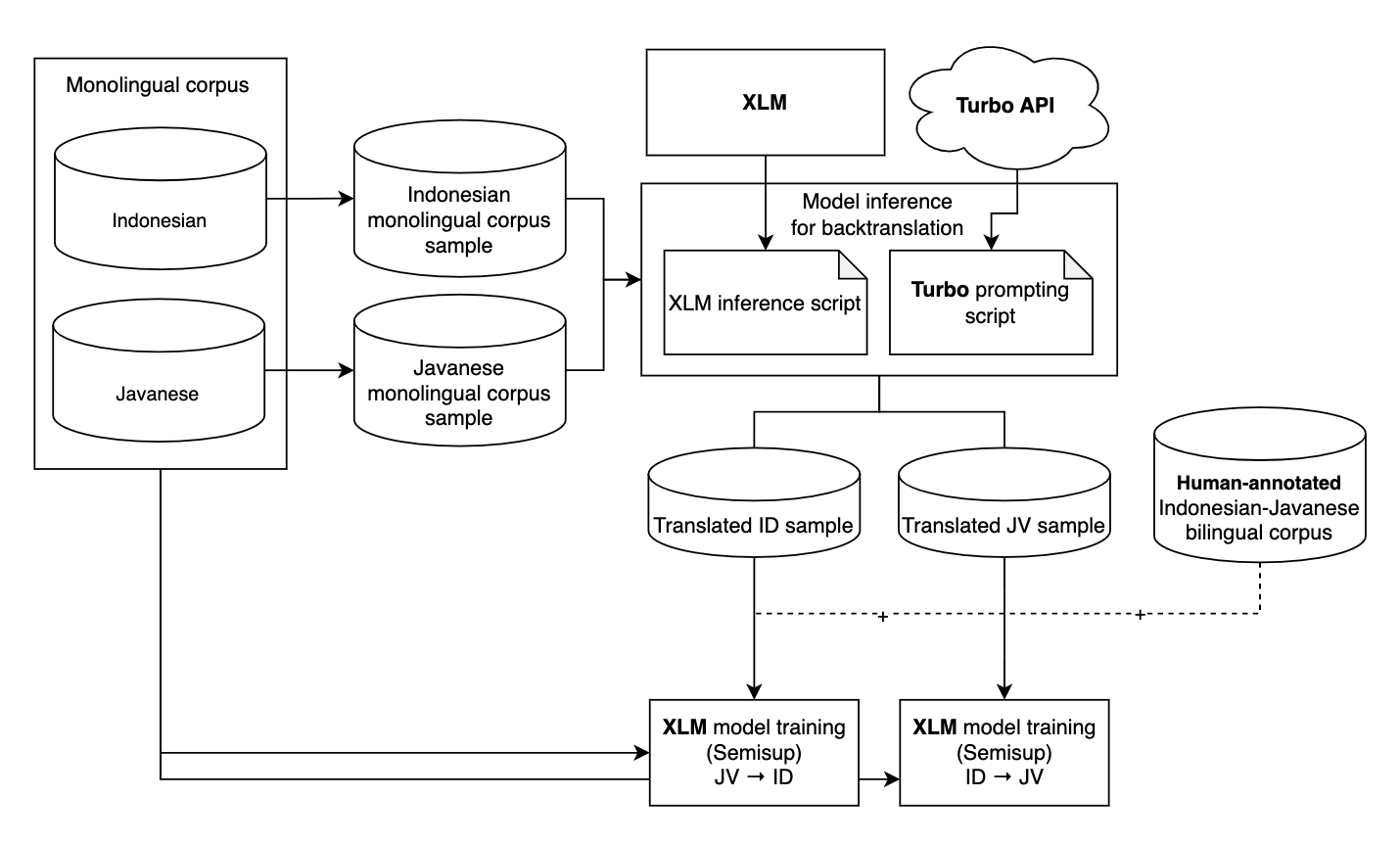}
    \caption{Illustration of translating monolingual data using an already trained model. In this work, we choose gpt-3.5-turbo and our NMT model trained on id-jv language pair using \textbf{Scratch} approach and \textbf{Semisup} paradigm.}
    \label{fig:bt_illustration}
\end{figure}

\begin{table}[ht]
\small
    \centering
    \begin{tabular}{lrr}
    \hline
        \textbf{Method} & \textbf{id$\rightarrow$jv}  & \textbf{jv$\rightarrow$id} \\
    \hline
        \verb|BT (baseline XLM)| & \verb|10.35| & \verb|12.29|\\
        \verb|BT (gpt-3.5)|      & \verb|13.22| & \verb|13.80|\\
    \hline
    \end{tabular}
    \caption{Comparison of additional data generation techniques. BT is for Back translation, in which we sample sentences from external monolingual corpora and translate them using the model indicated in the parentheses.}
    \label{tab:bt-exploratory}
\end{table} 

Besides prompting gpt-3.5 to generate parallel sentences directly, we also compared it with generating additional data from translating existing monolingual datasets. We use gpt-3.5 and the baseline XLM model to translate Wikipedia monolingual sentences. gpt-3.5-turbo is used instead of text-davinci-003 based on the results of the experiments shown in table \ref{tab:llm-idjv}. Our findings show that additional data from translating monolingual corpus using the baseline XLM model does not yield any significant performance increase or even hurts it, as shown in Table \ref{tab:bt-exploratory}, whereas monolingual corpus translated using gpt-3.5 yields over 1 BLEU score on the Javanese to Indonesian translation direction.

However, this increase is modest compared to the results shown by directly generating parallel sentences from gpt-3.5-turbo as additional parallel data. Therefore, we move forward with approach (1). We apply the same procedure to the remaining language pairs: Sundanese, Minangkabau, and Balinese. Synthetic data generation is a promising research avenue in which both approaches (1) and (2) should still be included.

\section{Hyperparameters}
\label{sec:hyperparams}
In this work, there are a total of ten combinations of training approaches and paradigms which are:
\begin{itemize}[noitemsep,topsep=0pt]
    \item \textbf{Scratch} - \textbf{Unsup}
    \item \textbf{Scratch} - \textbf{Semisup}
    \item \textbf{Scratch$_{AUG}$} - \textbf{Semisup}
    \item \textbf{PreXL} - Pre-training Language Model
    \item \textbf{PreXL} - \textbf{Unsup}
    \item \textbf{PreXL} - \textbf{Semisup}
    \item \textbf{CodeXL} - Pre-trained Language Model
    \item \textbf{CodeXL} - \textbf{Unsup}
    \item \textbf{CodeXL} - \textbf{Semisup}
    \item \textbf{CodeXL$_{AUG}$} - Semisup
\end{itemize}

Across all of these models, nine hyperparameters are shared among them:
\begin{itemize}[noitemsep, topsep=0pt]
    \item \texttt{emb\_dim}: 1024
    \item \texttt{n\_layers}: 6
    \item \texttt{n\_heads}: 8
    \item \texttt{dropout}: 0.1
    \item \texttt{attention\_dropout}: 0.1
    \item \texttt{gelu\_activation}: true
    \item \texttt{batch\_size}: 32
    \item \texttt{bptt}: 256
    \item \texttt{epoch\_size}: 200000
\end{itemize}

Models that are training using the \textbf{Unsup} training paradigm sets the hyperparameter \textbf{lambda\_ae} to \textbf{\emph{0:1,100000:0.1,300000:0}}. 

The hyperparameter \textbf{max\_vocab} is set to \textbf{\emph{200000}} for all models trained except models in \textbf{CodeXL} - Pre-trained Language Model, which uses the default values.

The hyperparameter \textbf{optimizer} is set to \textbf{\emph{adam\_inverse\_sqrt, beta1=0.9, beta2=0.98, lr=0.0001}} for all models trained except models in \textbf{CodeXL} - Pre-trained Language Model and "\textbf{CodeXL} - \textbf{Unsup}" which set the value to \textbf{\emph{adam, lr=0.0001}}.

The hyperparameter \textbf{tokens\_per\_batch} is set to \textbf{\emph{2000}} for all models excluding all models in \textbf{CodeXL} and \textbf{Scratch$_{AUG}$} which set the value to \textbf{\emph{4000}}.






\end{document}